\documentclass[runningheads]{llncs}
\usepackage{graphicx}
\usepackage{adjustbox}
\usepackage{algorithm,algpseudocode}
\newtheorem{exmp}{Example}
\usepackage{tabularx}
\algnewcommand\algorithmicinput{\textbf{Input:}}
\algnewcommand\Input{\item[\algorithmicinput]}
\algnewcommand\algorithmicoutput{\textbf{Output:}}
\algnewcommand\Output{\item[\algorithmicoutput]}
\algnewcommand\algorithmicforeach{\textbf{for each}}
\algnewcommand{\LeftComment}[1]{\Statex \(\triangleright\) #1}
\algdef{S}[FOR]{ForEach}[1]{\algorithmicforeach\ #1\ \algorithmicdo}

\usepackage{array}
\newcolumntype{C}[1]{>{\centering\arraybackslash}p{#1}}

\begin{document}
\title{FOLD-R++: A Scalable Toolset for Automated Inductive Learning of Default Theories from Mixed Data}
\titlerunning{FOLD-R++ Toolset}
%
\author{Huaduo Wang and Gopal Gupta }
\authorrunning{Wang and Gupta}
%
\institute{Department of Computer Science\\The University of Texas at Dallas, USA\\
\email{\{huaduo.wang, gupta\}@utdallas.edu}
}
\maketitle              
\begin{abstract}
FOLD-R is an automated inductive learning algorithm for learning default rules for mixed (numerical and categorical) data. It generates an (explainable) normal logic program (NLP) rule set for classification tasks. We present an improved FOLD-R algorithm, called FOLD-R++, that significantly increases the efficiency and scalability of FOLD-R by orders of magnitude. FOLD-R++ improves upon FOLD-R without compromising or losing information in the input training data during the encoding or feature selection phase. The FOLD-R++ algorithm is competitive in performance with the widely-used XGBoost algorithm, however, unlike XGBoost, the FOLD-R++ algorithm produces an explainable model. FOLD-R++ is also competitive in performance with the RIPPER system, however, on large datasets FOLD-R++ outperforms RIPPER. We also create a powerful tool-set by combining FOLD-R++ with s(CASP)---a goal-directed \textit{answer set programming (ASP)} execution engine---to make predictions on new data samples using the normal logic program generated by FOLD-R++. The s(CASP) system also produces a justification for the prediction. Experiments presented in this paper show that our improved FOLD-R++ algorithm is a significant improvement over the original design and that the s(CASP) system can make predictions in an efficient manner as well.

\end{abstract}

\section{Introduction}
Dramatic success of machine learning has led to a torrent of Artificial Intelligence (AI) applications. However, the effectiveness of these systems is limited by the machines' current inability to explain their decisions and actions to human users. That's mainly because the statistical machine learning methods produce models that are complex algebraic solutions to optimization problems such as risk minimization
or geometric margin maximization. Lack of intuitive descriptions makes it hard for users to understand and verify the underlying rules that govern the model. Also, these methods cannot produce a justification for a prediction they arrive at for a new data sample.
The Explainable AI program \cite{xai} aims to create a suite of machine learning techniques that: a) Produce more explainable models, while maintaining a high level of prediction accuracy; and b) Enable human users to understand, appropriately trust, and effectively manage the emerging generation of artificially intelligent systems. Inductive Logic Programming (ILP) \cite{ilp} is one Machine Learning technique where the learned model is in the form of logic programming rules that are comprehensible to humans. It allows the background knowledge to be incrementally extended without requiring the entire model to be re-learned. Meanwhile, the comprehensibility of symbolic rules makes it easier for users to understand and verify induced models and even refine them.

The ILP learning problem can be regarded as a search problem for a set of clauses that deduce the training examples. The search is performed either top down or bottom-up. A bottom-up approach builds most-specific clauses from the training examples and searches the hypothesis space by using generalization. This approach is not applicable to large-scale datasets, nor it can incorporate \textit{negation-as-failure} into the hypotheses. A survey of bottom-up ILP systems and their shortcomings can be found at \cite{sakama05}. In contrast, the top-down approach starts with the most general clause and then specializes it. A top-down algorithm guided by heuristics is better suited for large-scale and/or noisy datasets \cite{quickfoil}.

The FOIL algorithm \cite{foil} by Quinlan is a popular top-down inductive logic programming algorithm that generates logic programs. FOIL uses weighted information gain (IG) as the heuristics to guide the search for best literals. The FOLD algorithm by Shakerin \cite{shakerin-phd,fold} is a  new top-down algorithm inspired by the FOIL algorithm. It generalizes the FOIL algorithm by learning \textit{default rules with exceptions}. It does so by first learning the default predicate that covers positive examples while avoiding negative examples, then next it swaps the positive and negative examples and calls itself recursively to learn the exception to the default. Both FOIL and FOLD cannot deal with numeric features directly; an encoding process is needed in the preparation phase of the training data that discretizes the continuous numbers into intervals. However, this process not only adds a huge computational overhead to the algorithm but also leads to loss of information in the training data.

To deal with the above problems, Shakerin developed an extension of the FOLD algorithm, called FOLD-R, to handle mixed (i.e., both numerical and categorical) features which avoids the discretization process for numerical data \cite{shakerin-phd,fold}. However, FOLD-R still suffers from efficiency and scalability issues when compared to other popular machine learning systems for classification. 
In this paper we report on a novel implementation method we have developed to improve the design of the FOLD-R system. 
In particular, we use the prefix sum technique \cite{prefixsum} to optimize the process of calculation of information gain, the most time consuming component of the FOLD family of algorithms \cite{shakerin-phd}. Our optimization, in fact, reduces the time complexity of the algorithm. If $N$ is the number of unique values from a specific feature and $M$ is the number of training examples, then the complexity of computing information gain for all the possible literals of a feature is reduced from ${O(M*N)}$ for FOLD-R to $O(M)$ in FOLD-R++.


%
%
In addition to using prefix sum, we also improved the FOLD-R algorithm by allowing negated literals in the default portion of the learned rules (explained later). Finally, a hyper-parameter, called \textit{exception ratio}, which controls the training process that learns exception rules, is also introduced. This hyper-parameter helps improve efficiency and classification performance. These three changes make FOLD-R++ significantly better than FOLD-R and competitive with well-known algorithms such as XGBoost and RIPPER. 

Our experimental results indicate that the FOLD-R++ algorithm is comparable to popular machine learning algorithms such as XGBoost \cite{xgboost} and RIPPER \cite{ripper} wrt various metrics (accuracy, recall, precision, and F1-score) as well as in efficiency and scalability. However, in addition, FOLD-R++ produces an explainable and interpretable model in the form of a normal logic program. A normal logic program is a logic program extended with negation-as-failure \cite{lloyd}. Note that RIPPER also generates a set of CNF formulas to explain the model, however, as we will see later, FOLD-R++ outperforms RIPPER on large datasets.

This paper makes the following novel contribution: it presents the FOLD-R++ algorithm that significantly improves the efficiency and scalability of the FOLD-R ILP algorithm without adding overhead during pre-processing or losing information in the training data. 
As mentioned, the new approach is competitive with popular classification models such as the XGBoost classifier \cite{xgboost} and the RIPPER system \cite{ripper}. 
The FOLD-R++ algorithm outputs a normal logic program (NLP) \cite{lloyd,gelfondkahl} that serves as an explainable/interpretable model. This generated normal logic program is compatible with s(CASP) \cite{scasp}, a goal-directed ASP solver, that can efficiently justify the prediction generated by the ASP model.\footnote{The s(CASP) system is freely available at \url{https://gitlab.software.imdea.org/ciao-lang/sCASP}.}

\section{Background}
\label{sec:background}

\subsection{Inductive Logic Programming}
Inductive Logic Programming (ILP) \cite{ilp} is a subfield of machine learning that learns models in the form of logic programming rules that are comprehensible to humans. This problem is formally defined as:
\medskip 

\noindent 
\textbf{Given}
\begin{enumerate}
    \item A background theory $B$, in the form of an extended logic program, i.e., clauses of the form $h \leftarrow l_1, ... , l_m,\ not \ l_{m+1},...,\ not \ l_n$, where $l_1,...,l_n$ are positive literals and \textit{not} denotes \textit{negation-as-failure} (NAF) \cite{lloyd,gelfondkahl}. We require that $B$ has no  loops through negation, i.e., it is stratified \cite{lloyd}.
    \item Two disjoint sets of ground target predicates $E^+, E^-$ known as \emph{positive} and \emph{negative} examples, respectively
    \item A hypothesis language of function free predicates $L$, and a  refinement operator $\rho$ under $\theta$-subsumption \cite{plotkin70} that would disallow loops over negation.
\end{enumerate}
\textbf{Find} a set of clauses $H$ such that:
\begin{itemize}
    \item $ \forall e \in \ E^+ ,\  B \cup H \models e$
    \item $ \forall e \in \ E^- ,\  B \cup H \not \models e$
    \item $B \land H$ is consistent.
\end{itemize}

\subsection{Default Rules}

Default Logic \cite{reiter80,gelfondkahl} is a non-monotonic logic to formalize commonsense reasoning. A default $D$ is an expression of the form 

$$ A: \textbf{M} B \over\Gamma$$

\noindent which states that the conclusion $\Gamma$ can be inferred if pre-requisite $A$ holds and $B$ is justified. $\textbf{M} B$ stands for ``it is consistent to believe $B$" \cite{gelfondkahl}.
Normal logic programs can encode a default quite elegantly. A default of the form: 

$$\alpha_1 \land \alpha_2\land\dots\land\alpha_n: \textbf{M} \lnot \beta_1, \textbf{M} \lnot\beta_2\dots\textbf{M}\lnot\beta_m\over \gamma$$

\noindent can be formalized as the
following normal logic program rule:

$$\gamma ~\texttt{:-}~ \alpha_1, \alpha_2, \dots, \alpha_n, \texttt{not}~ \beta_1, \texttt{not}~ \beta_2, \dots, \texttt{not}~ \beta_m.$$

\noindent where $\alpha$'s and $\beta$'s are positive predicates and \texttt{not} represents negation-as-failure. We call such rules \emph{default rules}. 
Thus, the default $bird(X): M \lnot penguin(X)\over fly(X)$ will be represented as the following default rule in normal logic programming:

~~~~~~~~~{\tt fly(X) :- bird(X), not penguin(X).}

\noindent We call {\tt bird(X)}, the condition that allows us to jump to the default conclusion that {\tt X} can fly, the {\it default part} of the rule, and {\tt not penguin(X)} the \textit{exception part} of the rule. 

Default rules closely represent the human thought process (commonsense reasoning). FOLD-R and FOLD-R++ learn default rules represented as normal logic programs. An advantage of learning default rules is that we can distinguish between exceptions and noise \cite{fold,shakerin-phd}. Note that the programs currently generated by the FOLD-R++ system are stratified normal logic programs \cite{lloyd}.

\begin{algorithm}[!h]
\caption{FOLD-R Algorithm}
\label{algo:fold-r-original}
\begin{algorithmic}[1]
\Input $B$: background knowledge, $E^+$: positive example, $E^-$: negative example
\Output $D = \{c_1, ..., c_n\}$: a set of defaults rules with exceptions
\Function{Fold}{$E^+, E^-$} \Comment{$target, B$: global vars}
\State $D \gets$ \O
\While{$|E^+| > 0$}
\State $c \gets$ \Call{specialize}{{$target$ :- $ \ true$}, {$E^+$}, {{$E^-$}}} 
\State $E^+ \gets E^+ \setminus covers(c,\ E^+)$ \Comment{rule out already covered examples}
\State $D \gets D \cup \{ c \}$
\EndWhile 
\State \Return $D$
\EndFunction

\Function{Specialize}{${c},{E^+},{E^-}$}
\While{$|E^-|>0$}
\State $c',\ ig \gets$ \Call{add\_best\_literal}{{c},\ {$E^+$},\ {{$E^-$}}}
\If{$ig > 0$}
\State $c \gets c' $
\Else
\State $c \gets \Call{exception}{c,\ {E^+},\ {E^-}}$
\If {$c $ is $null$}
\State $c \gets enumerate(c,\ E^+)$ \Comment{generate clause to maximally cover $E^+$}
\EndIf	
\EndIf
\State $E^+ \gets covers(c,\ E^+)$
\State $E^- \gets covers(c,\ E^-)$
\EndWhile
\State \Return $c$
\EndFunction

\Function{Exception}{${c}, {E^+}, {E^-}$} 
\State $ AB \gets \Call{fold}{E^-,\ E^+} $ \Comment{recursively call FOLD after swapping $E^+$ and $E^-$}
\If{$AB$ is \O}
\State $c \gets null$
\Else 
\State $c \gets set\_exception(c,\ AB)$ \Comment{set exception part of clause $c$ as $AB$}
\EndIf
\State \Return $c$
\EndFunction

\Function{Add\_best\_literal}{${c}, {E^+}, {E^-}$} 
\LeftComment{return the clause with the best literal added and its corresponding info gain}
    \State $c_1,\ ig_1 \gets best\_categorical(c,\ E^+,\ E^-)$
    \State $c_2,\ ig_2 \gets best\_numerical(c,\ E^+,\ E^-)$ \Comment{FOLD-R extension}
	\If{$c_1 > c_2$}
	\State \Return $c_1,\ ig_1$
    \Else
    \State \Return $c_2,\ ig_2$
    \EndIf

\EndFunction

\end{algorithmic}
\end{algorithm}

\section{The FOLD-R Algorithm} 

The FOLD algorithm \cite{shakerin-phd,fold} is a top-down ILP algorithm that searches for best literals to add to the body of the clauses for hypothesis, $H$, with the guidance of an information gain-based heuristic. 
The FOLD-R algorithm is a numeric extension of the FOLD algorithm that adopts the approach of the well-known C4.5 algorithm \cite{C45} for finding literals. 
Algorithm \ref{algo:fold-r-original} gives an overview of the FOLD-R algorithm. The extended algorithm will directly select the best numerical literal, in addition to selecting the categorical literals.
Thus, the best\_numerical function (line 37 in Algorithm \ref{algo:fold-r-original}) finds the best numerical literal and adds it to the clause after classifying all the training examples for each numerical split on all the features. The other functions remain the same as the FOLD algorithm \cite{fold,shakerin-phd}. We illustrate the FOLD-R algorithm through an example.

\begin{exmp}
\label{ex:pinguin}
In the FOLD-R algorithm, the target is to learn rules for \texttt{fly(X)}. $ B, E^+, E^-$ are background knowledge, positive and negative examples, respectively.
\end{exmp}
\begin{verbatim}
B:  bird(X) :- penguin(X).
    bird(tweety).   bird(et).
    cat(kitty).     penguin(polly).
E+: fly(tweety).    fly(et).
E-: fly(kitty).     fly(polly).
\end{verbatim}

The target predicate \texttt{\{fly(X) :- true.\}} is specified when calling the specialize function at line 4 in Algorithm \ref{algo:fold-r-original}. The add\_best\_literal function selects the literal \texttt{bird(X)} as a result and adds it to the clause $r$ = \texttt{fly(X) :- bird(X)} because it has the best information gain among \texttt{\{bird,penguin,cat\}} at line 12. Then, the training set gets updated to E$^+$=\texttt{\{tweety, et\}}, E$^-$=\texttt{\{polly\}} at line 21--22 in SPECIALIZE function.
The negative example \texttt{polly} is still falsely implied by the generated clause.
The default learning of SPECIALIZE function is finished because the information gain of candidate literal $c'$ is zero. 
Therefore, the exception learning starts by calling FOLD function recursively with swapped positive and negative examples, E$^+$=\texttt{\{polly\}}, E$^-$=\texttt{\{tweety, et\}} at line 27.
In this case, an abnormal predicate \texttt{\{ab0(X) :- penguin(X)\}} is generated and returned as the only exception to the previous learned clause as $r$ = \texttt{fly(X) :- bird(X), not ab0(X)}. The abnormal rule \texttt{\{ab0(X) :- penguin(X)\}} is added to the final rule set producing the program below:

\begin{center}
    \begin{tabular}{l}
        \texttt{fly(X) :- bird(X), not ab0(X).}\\     
        \texttt{ab0(X) :- penguin(X).}     
    \end{tabular}
\end{center}

\section{The FOLD-R++ Algorithm}

The FOLD-R++ algorithm  
refactors the FOLD-R algorithm. FOLD-R++ makes three main improvements to FOLD-R: (i) it can learn and add negated literals to the default (positive) part of the rule; in the FOLD-R algorithm negated literals can only be in the exception part, (ii) prefix sum algorithm is used to speed up computation, and (iii) a hyper parameter called $ratio$ is introduced to control the level of nesting of exceptions. These three improvements make FOLD-R significantly more efficient.

The FOLD-R++ algorithm is summarized in Algorithm \ref{algo:fold}. The output of the FOLD-R++ algorithm is a set of default rules \cite{gelfondkahl} coded as a normal logic program. An example implied by any rule in the set would be classified as positive.  Therefore, the FOLD-R++ algorithm rules out the already covered positive examples at line 9 after learning a new rule. To learn a particular rule, 
the best literal would be repeatedly selected---and added to the default part of the rule's body---based on information gain using the remaining training examples (line 17). 
Next, only the examples that can be covered by learned default literals would be used for further learning (specializing) of the current rule (line 20--21).
When the information gain becomes zero or the number of negative examples drops below the \textit{ratio} threshold, the learning of the default part is done. 
FOLD-R++ next learns exceptions after first learning default literals. This is done by swapping the residual positive and negative examples and calling itself recursively in line 26. The remaining positive and negative examples can be swapped again and exceptions to exceptions learned (and then swapped further to learn exceptions to exceptions of exceptions, and so on). The $ratio$ parameter in Algorithm \ref{algo:fold} represents the ratio of training examples that are part of the exception to the examples implied by only the default conclusion part of the rule. It allows users to control the nesting level of exceptions.

Generally, avoiding falsely covering negative examples by adding literals to the default part of a rule will reduce the number of positive examples the rule can imply. 
Explicitly activating the exception learning procedure (line 26) could increase the number of positive examples a rule can cover while reducing the total number of rules generated. As a result, the interpretability is increased due to fewer rules and literals being generated. For the Adult Census Income dataset, for example, without the hyper-parameter exception \textit{ratio} (equivalent to setting the \textit{ratio} to 0), the FOLD-R++ algorithm would take around 10 minutes to finish the training and generate hundreds of rules. With the $ratio$ parameter set to 0.5, only 13 rules are generated in around 10 seconds.

Additionally, The FOLD and FOLD-R algorithms disabled the negated literals in the default theories to make the generated rules look more elegant (only exceptions included negated literals). However, a negated literal sometimes is the optimal literal with the most useful information gain. FOLD-R++ allows for negated literals in the default part of the generated rules.
We cannot make sure that FOLD-R++ generates optimal combination of literals because it is a greedy algorithm, however, it is an improvement over FOLD and FOLD-R.

\begin{algorithm}[!h]
\caption{FOLD-R++ Algorithm}
\label{algo:fold}
\begin{algorithmic}[1]
\Input $E^+$: positive examples, $E^-$: negative examples
\LeftComment Global Parameters: $target$, $B$: background knowledge,  $ratio$: exception ratio 
\Output  $R = \{r_1,...,r_n\}$: a set of defaults rules with exceptions 
\Function{Fold\_rpp}{$E^+, E^-, L_{used}$} \Comment{$L_{used}$: used literals, initially empty}
\State $R \gets$ \O
\While{$|E^+| > 0$}
\State $r \gets$ \Call{learn\_rule}{{$E^+$}, {$E^-$}, {$L_{used}$}}
\State $E_{tp} \gets covers(r,\ E^+)$ \Comment{$E_{tp}$: true positive examples implied by rule $r$}
\If {$|E_{tp}|=0$}
\State \textbf{break}
\EndIf
\State $E^+ \gets E^+ \setminus\ E_{tp}$ \Comment{rule out the already covered examples}
\State $R \gets R\ \cup\ \{ r \}$
\EndWhile
\State \Return $R$
\EndFunction
\Function{Learn\_rule}{${E^+}, {E^-}, {L_{used}}$}
\State $L \gets$ \O \Comment{$L$: default literals for the result rule $r$}
\While{$true$}
\State  $l \gets$ \Call{find\_best\_literal}{{$E^+$}, {{$E^-$}}, {$L_{used}$}}
\State $L \gets L\ \cup\ \{ l \}$
\State $r \gets \textit{set\_default}(r,\ L)$ \Comment{set default part of rule $r$ as $L$}
\State $E^+ \gets covers(r,\ E^+)$ 
\State $E^- \gets covers(r,\ E^-)$
\If{$l$ is invalid or $|E^-| \leq |E^+| * ratio$}
\If{$l$ is invalid} 
\State $r \gets \textit{set\_default}(r,\ L\setminus\ \{ l \})$ \Comment{remove the invalid literal $l$ from rule $r$}
\Else   
\State $AB \gets$ \Call{fold\_rpp}{{$E^-$}, {{$E^+$}}, {$L_{used} + L$}} \Comment{learn exception rules for $r$}
\State $r \gets set\_exception(r, AB)$ \Comment{set exception part of rule $r$ as $AB$}
\EndIf
\State \textbf{break}
\EndIf
\EndWhile
\State \Return $r$ \Comment{the head of rule $r$ is $target$}
\EndFunction

\end{algorithmic}
\end{algorithm}

\subsection{Literal Selection}

The literal selection process for Shakerin's FOLD-R algorithm can be summarized as function SPECIALIZE in Algorithm \ref{algo:fold-r-original}.
The FOLD-R algorithm \cite{shakerin-phd,fold} selects the best literal based on the weighted information gain for learning defaults, similar to the original FOLD algorithm described in \cite{fold}. For numeric features, the FOLD-R algorithm would enumerate all the possible splits. Then, it classifies the data and computes information gain for literals for each split. The literal with the best information gain would be selected as a result. In contrast, the FOLD-R++ algorithm uses a new, more efficient method employing \textit{prefix sums} to calculate the information gain based on the classification categories. The FOLD-R++ algorithm divides features into two categories: \emph{categorical} and \emph{numerical}. All the values in a categorical feature would be considered as categorical values even if some of them are numbers. Only equality and inequality literals would be generated for categorical features. For numerical features, the FOLD-R++ algorithm would try to read each value as a number, converting it to a categorical value if the conversion fails. Additional numerical comparison ($\leq$ and $>$) literal candidates would be generated for numerical features. A mixed type feature that contains both categorical and numerical values would be treated as a numerical feature.

In FOLD-R++, information gain for a given literal is calculated as shown in Algorithm \ref{algo:fold-r-plus-ig}. The variables $tp, fn, tn, fp$ for finding the information gain represent the numbers of true positive, false negative, true negative, and false positive examples, respectively. With the simplified information gain function $IG$ in Algorithm \ref{algo:fold-r-plus-ig}, the new approach employs the \textit{prefix sum technique} to speed up the calculation. Only one round of classification is needed for a single feature, even with mixed types of values.

\begin{algorithm}[!h]
\caption{FOLD-R++ Algorithm, Information Gain function}

\label{algo:fold-r-plus-ig}
\begin{algorithmic}[1]
\Input $tp$, $fn$, $tn$, $fp$: the number of E$_{tp}$, E$_{fn}$, E$_{tn}$, E$_{fp}$ implied by literal
\Output  information gain 
\Function{IG}{$tp,fn,tn,fp$} \Comment{IG is the function that computes information gain}
\If{$fp+fn>tp+tn$}
\State \Return $-\infty$
\EndIf
\State \Return $\frac{1}{tp+fp+tn+fn} \cdot $(\Call{F}{$tp,fp$} + \Call{F}{$fp,tp$} + \Call{F}{$tn,fn$} + \Call{F}{$fn,tn$})
\EndFunction

\Function{F}{$a,b$}
\If{$a=0$}
\State \Return 0
\EndIf
\State \Return $a \cdot \textrm{log}_2(\frac{a}{a+b})$
\EndFunction

\end{algorithmic}
\end{algorithm}

\begin{algorithm}[!h]
\caption{FOLD-R++ Algorithm, Find Best Literal function}
\label{algo:best_ig}
\begin{algorithmic}[1]
\Input $E^+$: positive examples, $E^-$: negative examples, $L_{used}$: used literals
\Output  $best\_lit$: the best literal that provides the most information 

\Function{Find\_best\_literal}{$E^+,E^-,L_{used}$} 
\State $best\_ig, best\_lit \gets -\infty, invalid$
\For{$i \gets 1$  to  $N$} \Comment{$N$ is the number of features}
\State $ig, lit \gets$ \Call{best\_info\_gain}{{$E^+$}, {$E^-$}, {i}, {$L_{used}$}}
\If {$best\_ig < ig$}
\State $best\_ig, best\_lit \gets ig, lit$
\EndIf
\EndFor
\State \Return $best\_lit$ 
\EndFunction

\Function{Best\_info\_gain}{$E^+,E^-,i,L_{used}$} \Comment{$i$: feature index}
\State $pos, neg \gets count\_classification(E^+, E^-, i)$
\LeftComment{$pos$ ($neg$): dicts that holds the numbers of $E^+$ ($E^-$) for each unique value}
\State $xs, cs \gets collect\_unique\_values(E^+, E^-, i)$ 
\LeftComment{$xs$ ($cs$): lists that holds the unique numerical (categorical) values}
\State $xp, xn, cp, cn \gets count\_total(E+, E-, i)$  
\LeftComment{$xp$ ($xn$): the total number of $E^+$ ($E^-$) with numerical value.}
\LeftComment{$cp$ ($cn$): the total number of $E^+$ ($E^-$) with categorical value.}
\State $xs \gets couting\_sort(xs)$ 
\For{$j \gets 1$  to  $size(xs)$} \Comment{compute prefix sum for $E^+$ \& $E^-$ numerical values}
\State $pos[xs_{i}] \gets pos[xs_{i}] + pos[xs_{i-1}]$
\State $neg[xs_{i}] \gets neg[xs_{i}] + neg[xs_{i-1}]$
\EndFor

\For{$x \in xs$} \Comment{compute info gain for numerical comparison literals}
\State $lit\_dict[literal(i,\leq, x)] \gets$ \Call{IG}{$pos[x], xp-pos[x]+cp,xn-neg[x]+cn,neg[x]$}
\State $lit\_dict[literal(i,>, x)] \gets$ \Call{IG}{$xp-pos[x],pos[x]+cp,neg[x]+cn,xn-neg[x]$}
\EndFor

\For{$c \in cs$} \Comment{compute info gain for equality comparison literals}
\State $lit\_dict[literal(i,=, x)] \gets$ \Call{IG}{$pos[c],cp-pos[c]+xp,cn-neg[c]+xn,neg[c]$}
\State $lit\_dict[literal(i,\ne, x)] \gets$ \Call{IG}{$cp-pos[c]+xp,pos[c],neg[c],cn-neg[c]+xn$}
\EndFor
\State $best\_ig, lit \gets best\_pair(lit\_dict, L_{used})$ 
\State \Return $best\_ig, lit$ \Comment{return the best info gain and its corresponding literal}
\EndFunction

\end{algorithmic}
\end{algorithm}

In the FOLD-R++ algorithm, two types of literals would be generated: \emph{equality comparison} literals and \emph{numerical comparison} literals. The equality (\textit{resp.} inequality) comparison is straightforward in FOLD-R++: two values are equal if they are same type and identical, else they are unequal. However, a different assumption is made for comparisons between a numerical value and categorical value in FOLD-R++. Numerical comparisons ($\leq$ and $>$) between a numerical value and a categorical value is always false. A comparison example is shown in Table \ref{tab:comparison} (Left), while an evaluation example for a given literal, $literal{(i,\leq,3)}$, based on the comparison assumption is shown in Table \ref{tab:comparison} (Right). Given E$^+=\{1,2,3,3,5,6,6,b\}$, E$^-=\{2,4,6,7,a\}$, and $literal{(i,\leq,3)}$, the true positive example E$_{tp}$, false negative examples E$_{fn}$, true negative examples E$_{tn}$, and false positive examples E$_{fp}$ implied by the literal are $\{1,2,3,3\}$, $\{5,6,6,b\}$, $\{4,6,7,a\}$, $\{2\}$ respectively. Then, the information gain of literal$(i,\leq,3)$ is calculated $IG_{(i,\leq,3)}(4,4,4,1)=-0.619$ through Algorithm \ref{algo:fold-r-plus-ig}.

\begin{table}[]
\centering
\begin{tabular}{cc}
    \begin{minipage}{.4\linewidth}
        \centering
        \begin{tabular}{|c|c|}
        \cline{1-2}
        \textbf{comparison}  & \textbf{evaluation} \\
        \cline{1-2}
        3 $=$ `a' & False \\ 
        \cline{1-2}
        3 $\neq$ `a' & True \\ 
        \cline{1-2}
        3 $\le$ `a' & False \\ 
        \cline{1-2}
        3 $>$ `a' & False \\ 
        \cline{1-2}
        \end{tabular}
    \end{minipage} &

    \begin{minipage}{.55\linewidth}
        \centering
        
        \begin{tabular}{|c|c|c|}
        \cline{1-3}
          & \multicolumn{1}{c|}{\textbf{i$^{th}$ feature values}} & \textbf{count} \\
        \cline{1-3}
        $\mathbf{E^+}$ & 1 2 3 3 5 6 6 b & \textbf{8} \\
        \cline{1-3}
        $\mathbf{E^-}$ & 2 4 6 7 a & \textbf{5} \\
        \cline{1-3}
        $\mathbf{E_{tp(i,\leq,3)}}$ & 1 2 3 3 & \textbf{4} \\
        \cline{1-3}
        $\mathbf{E_{fn(i,\leq,3)}}$ & 5 6 6 b & \textbf{4} \\
        \cline{1-3}
        $\mathbf{E_{tn(i,\leq,3)}}$ & 4 6 7 a & \textbf{4} \\
        \cline{1-3}
        $\mathbf{E_{fp(i,\leq,3)}}$ & 2 & \textbf{1} \\
        \cline{1-3}
        \end{tabular}
    \end{minipage} 
\end{tabular}
\caption{Left: Comparisons between a numerical value and a categorical value. Right: Evaluation and count for literal$(i,\leq,3)$. }
\label{tab:comparison}
\end{table}

The new approach to find the best literal that provides most useful information is summarized in Algorithm \ref{algo:best_ig}. 
In line 12, $pos$ ($neg$) is the dictionary that holds the numbers of positive (negative) examples for each unique value. 
In line 13, $xs$ ($cs$) is the list that holds the unique numerical (categorical) values.
In line 14, $xp$ ($xn$) is the total number of positive (negative) examples with numerical values; $cp$ ($cn$) is the total number of positive (negative) examples with categorical values.
After computing the prefix sum at line 16, $pos[x]$ ($neg[x]$) holds the total number of positive (negative) examples that have a value less than or equal to $x$.
Therefore, $xp-pos[x]$ ($xn-neg[x]$) represents the total number of positive (negative) examples that have a value greater than $x$.
In line 21, the information gain of literal$(i,\le,x)$ is calculated by calling Algorithm \ref{algo:fold-r-plus-ig}.
Note that  
$pos[x]$ ($neg[x]$) is the actual value for the formal parameter $tp$ ($fp$) of function IG in Algorithm \ref{algo:fold-r-plus-ig}. 
Likewise, $xp-pos[x]+cp$ ($xn-neg[x]+cn$) substitute for formal parameter $fn$ ($tn$) of the function IG. 
$cp$ ($cn$) is included in the actual parameter for formal parameter $fn$ ($tn$) of function IG because of the assumption that any numerical comparison between a numerical value and a categorical value is false.
The information gain calculation processes of other literals also follow the comparison assumption mentioned above. 
Finally, the best\_info\_gain function (Algorithm \ref{algo:best_ig}) returns the best score on information gain and the corresponding literal except the literals that have been used in current rule-learning process. 
For each feature, we compute the best literal, then the find\_best\_literal function returns the best literal among this set of best literals.
FOLD-R algorithm selects only positive literals in default part of rules during literal selection even if a negative literal provides better information gain. 
Unlike FOLD-R, the FOLD-R++ algorithm can also select negated literals for the default part of a rule at line 26 in Algorithm \ref{algo:best_ig}.


It is easy to justify the  ${O(M)}$ complexity of information gain calculation in FOLD-R++ mentioned earlier. The time complexity of Algorithm \ref{algo:fold-r-plus-ig} is obviously $O(1)$. Algorithm \ref{algo:fold-r-plus-ig} is called in line 21, 22, 25, and 26 of Algorithm \ref{algo:best_ig}. Line 12--15 in Algorithm \ref{algo:best_ig} can be considered as the preparation process for calculating information gain and has complexity $O(M)$, assuming that we use counting sort (complexity $O(M)$) with a pre-sorted list in line 15; it is easy to see that lines 16--29 take time $O(N)$.

\begin{exmp}
\label{ex:pinguin2}
Given positive and negative examples, $E^+, E^-$, with mixed type of values on feature $i$, the target is to find the literal with the best information gain on the given feature. There are $8$ positive examples, their values on feature $i$ are $[1,2,3,3,5,6,6,b]$. And, the values on feature $i$ of the $5$ negative examples are $[2,4,6,7,a]$. 

\end{exmp}

\noindent 
With the given examples and specified feature, the numbers of positive examples and negative examples for each unique value are counted first, which are shown as $pos, neg$ at right side of Table \ref{tbl:example1}. Then, the prefix sum arrays are calculated for computing the heuristic as psum$^+$, psum$^-$. Table \ref{tbl:example2} shows the information gain for each literal, the $literal(i, \ne, a)$ has been selected with the highest score.

\begin{table}[]
\centering
\begin{tabular}{|c|c|}
\hline
 & \textbf{i$^{th}$ feature values}  \\ \hline
$\mathbf{E^+}$                              & 1 2 3 3 5 6 6 b \\ \hline
$\mathbf{E^-}$                              & 2 4 6 7 a       \\ \hline
\end{tabular}
\quad
\begin{tabular}{|l|C{0.5cm}|C{0.5cm}|C{0.5cm}|C{0.5cm}|C{0.5cm}|C{0.5cm}|C{0.5cm}|C{0.5cm}|C{0.5cm}|}
\hline
\textbf{value}          & 1 & 2 & 3 & 4 & 5 & 6 & 7 & a  & b  \\ \hline
\textbf{pos}     		& 1 & 1 & 2 & 0 & 1 & 2 & 0 & 0  & 1  \\ \hline
\textbf{psum$^+$}    & 1 & 2 & 4 & 4 & 5 & 7 & 7 & na & na \\ \hline
\textbf{neg}   			& 0 & 1 & 0 & 1 & 0 & 1 & 1 & 1  & 0  \\ \hline
\textbf{psum$^-$}    & 0 & 1 & 1 & 2 & 2 & 3 & 4 & na & na \\ \hline
\end{tabular}
\caption{Left: Examples and values on i$^{th}$ feature. Right: positive/negative count and prefix sum on each value }
\label{tbl:example1}
\end{table}

\begin{table}[]
\centering
\begin{tabular}{|l|c|c|c|c|c|c|c|c|c|}
\hline
                    & \multicolumn{9}{c|}{Info Gain}  \\ \hline
\textbf{value}      & 1       & 2       & 3       & 4       & 5       & 6       & 7       & a        & b        \\ \hline
\textbf{$\leq$ value} & $-\infty$ & $-\infty$ & -0.619  & -0.661  & -0.642  & -0.616  & -0.661  & na       & na       \\ \hline
\textbf{$>$ value}  & -0.664  & -0.666  & $-\infty$ & $-\infty$ & $-\infty$ & $-\infty$ & $-\infty$ & na       & na       \\ \hline
\textbf{$=$ value} & na      & na      & na      & na      & na      & na      & na      & $-\infty$  & $-\infty$  \\ \hline
\textbf{$\ne$ value} & na      & na      & na      & na      & na      & na      & na      & \textbf{-0.588}  & -0.627  \\ \hline

\end{tabular}
\caption{The info gain on i$^{th}$ feature with given examples}
\label{tbl:example2}
\end{table}

\subsection{Explainability}

Explainability is very important for some tasks like loan approval, credit card approval, and disease diagnosis system. Inductive logic programming provides explicit rules for how a prediction is generated compared to black box models like those based on neural networks. To efficiently justify the prediction, the FOLD-R++ outputs normal logic programs that are compatible with the s(CASP) goal-directed answer set programming system \cite{scasp}. The s(CASP) system executes answer set programs in a goal-directed manner \cite{scasp}. Stratified normal logic programs output by FOLD-R++ are a special case of answer set programs.

\begin{exmp}
\label{ex:adult}
The ``Adult Census Income" is a classical classification task that contains 32561 records. We treat 80\% of the data as training examples and 20\% as testing examples. The task is to learn the income status of individuals (more/less than 50K/year) based on features such as gender, age, education, marital status, etc. FOLD-R++ generates the following program that contains only 13 rules: 
\end{exmp}

{\scriptsize
\begin{verbatim}
(1) income(X,'=<50k') :- not marital_status(X,'married-civ-spouse'), not ab4(X), not ab5(X). 
(2) income(X,'=<50k') :- education_num(X,N4), N4=<12.0, capital_gain(X,N10), N10=<5013.0, 
                         not ab6(X), not ab8(X). 
(3) income(X,'=<50k') :- occupation(X,'farming-fishing'), age(X,N0), N0>62.0, N0=<63.0, 
                         education_num(X,N4), N4>12.0, capital_gain(X,N10), N10>5013.0. 
(4) income(X,'=<50k') :- age(X,N0), N0>65.0, education_num(X,N4), N4>12.0, 
                         capital_gain(X,N10), N10>9386.0, N10=<10566.0. 
(5) income(X,'=<50k') :- age(X,N0), N0>35.0, fnlwgt(X,N2), N2>199136.0, education_num(X,N4), 
                         N4>12.0, capital_gain(X,N10), N10>5013.0, hours_per_week(X,N12), 
                         N12=<20.0. 
(6)  ab1(X) :- age(X,N0), N0=<20.0. 
(7)  ab2(X) :- education_num(X,N4), N4=<10.0, capital_gain(X,N10), N10=<7978.0. 
(8)  ab3(X) :- capital_gain(X,N10), N10>27828.0, N10=<34095.0. 
(9)  ab4(X) :- capital_gain(X,N10), N10>6849.0, not ab1(X), not ab2(X), not ab3(X). 
(10) ab5(X) :- age(X,N0), N0=<27.0, education_num(X,N4), N4>12.0, capital_loss(X,N11), 
               N11>1974.0, N11=<2258.0. 
(11) ab6(X) :- not marital_status(X,'married-civ-spouse'). 
(12) ab7(X) :- occupation(X,'transport-moving'), age(X,N0), N0>39.0. 
(13) ab8(X) :- education_num(X,N4), N4=<8.0, capital_loss(X,N11), N11>1672.0, N11=<1977.0, 
               not ab7(X). 
\end{verbatim}  
}

\noindent The above program achieves 0.86 accuracy, 0.88 precision, 0.95 recall, and 0.91 $F_1$ score. Given a new data sample, the predicted answer for this data sample using the above logic program can be efficiently produced by the s(CASP) system \cite{scasp}. Since s(CASP) is query driven, an example query such as {\tt ?- income(30, Y)} which checks the income status of the person with ID 30, will succeed if the income is indeed predicted as less equal to 50K by the model represented by the logic program above. 

The s(CASP) system will also produce a justification (a proof tree) for this prediction query. It can even generate this proof tree in English, i.e., in a more human understandable form \cite{arias-justification}. The justification tree generated for the person with ID 30  is shown below:

{\scriptsize
\begin{verbatim}
?- income(30,Y).
% QUERY:I would like to know if
     `income' holds (for 30, and Y).
ANSWER:	1 (in 2.246 ms)
JUSTIFICATION_TREE:
`income' holds (for 30, and `=<50k'), because
    there is no evidence that `marital_status' holds (for 30, and married-civ-spouse), and
    there is no evidence that `ab4' holds (for 30), because
        there is no evidence that `capital_gain' holds (for 30, and Var1), 
        with Var1 not equal 0.0, and `capital_gain' holds (for 30, and 0.0).
    there is no evidence that `ab5' holds (for 30), because
        there is no evidence that `age' holds (for 30, and Var2), with Var2 not equal 18.0, 
        and `age' holds (for 30, and 18.0), and
        there is no evidence that `education_num' holds (for 30, and Var3), 
        with Var3 not equal 7.0, and `age' holds (for 30, and 18.0), justified above, and
        `education_num' holds (for 30, and 7.0).
The global constraints hold.
BINDINGS:
Y equal `=<50k'
\end{verbatim}  
}

With the justification tree, the reason for the prediction can be easily understood by human beings. The generated NLP rule-set can also be understood by a human.  If there is any unreasonable logic generated in the rule set, it can also be modified directly by the human without retraining. Thus, any bias in the data that is captured in the generated NLP rules can be corrected by the human user, and the updated NLP rule-set used for making new predictions.

The RIPPER system \cite{ripper} is a well-known rule-induction algorithm that generates formulas in conjunctive normal form (CNF) as an explanation of the model. RIPPER generates 53 formulas for Example \ref{ex:adult} and achieves 0.61 accuracy, 0.98 precision, 0.50 recall, and 0.66 $F_1$ score. A few of the fifty three rules generated by RIPPER for this dataset are shown below.

{\scriptsize
\begin{verbatim}
(1)  marital_status=Never-married & education_num=7.0-9.0 & workclass=Private &
     hours_per_week=35.0-40.0 & capital_gain=<9999.9 & sex=Female
(2)  marital_status=Never-married & capital_gain=<9999.9 & education_num=7.0-9.0 &
     hours_per_week=35.0-40.0 & relationship=Own-child
(3)  marital_status=Never-married & capital_gain=<9999.9 & education_num=7.0-9.0 &
     hours_per_week=35.0-40.0 & race=White & age=22.0-26.0
(4)  marital_status=Never-married & capital_gain=<9999.9 & education_num=7.0-9.0 &
     hours_per_week=24.0-35.0
     ... ...
(50) education_num=7.0-9.0 & age=26.0-30.0 & fnlwgt=177927.0-196123.0 & workclass=Private
(51) relationship=Not-in-family & capital_gain=<9999.9 & hours_per_week=35.0-40.0 &
     sex=Female & education=Assoc-voc
(52) education_num=<7.0 & workclass=Private & fnlwgt=260549.8-329055.0
(53) relationship=Not-in-family & capital_gain=<9999.9 & hours_per_week=35.0-40.0 &
     education_num=11.0-13.0 & occupation=Adm-clerical
\end{verbatim}  
}

Generally, a set of default rules is a more succinct description of a given concept compared to a set of CNFs, especially when nested exceptions are allowed in the default rules. For this reason, we believe that FOLD-R++ performs better than RIPPER on large datasets, as shown later.

\section{Experiments}
\label{sec:experiments}

In this section, we present our experiments on UCI standard benchmarks \cite{uci}.\footnote{The FOLD-R++ system  is available at \url{https://github.com/hwd404/FOLD-R-PP.}} The XGBoost Classifier is a popular classification model and used as a baseline in our experiment. We used simple settings for XGBoost classifier without limiting its performance. However, XGBoost cannot deal with mixed type (numerical and categorical) of examples directly. One-hot encoding has been used for data preparation. We use precision, recall, accuracy, $F_1$ score, and execution time to compare the results.

FOLD-R++ does not require any encoding before training. We implemented FOLD-R++ with Python (the original FOLD-R implementation is in Java). 
To make inferences using the generated rules, we developed a simple logic programming interpreter for our application that is part of the FOLD-R++ system. 
Note that the generated programs are stratified, so implementing an interpreter for such a restricted class in Python is relatively easy. However, for obtaining the justification/proof tree, or for translating the NLP rules into equivalent English text, one must use the s(CASP) system.

The time complexity for computing information gain on a feature is significantly reduced in FOLD-R++ due to the use of prefix-sum, resulting in rather large improvements in efficiency. For the credit-a dataset with only 690 instances, the new FOLD-R++ algorithm is a hundred times faster than the original FOLD-R. The hyper-parameter ratio is simply set as 0.5 for all the experiments. All the learning experiments have been conducted on a desktop with Intel i5-10400 CPU @ 2.9GHz and 32 GB ram. To measure performance metrics, we conducted 10-fold cross-validation on each dataset and the average of accuracy, precision, recall, $F_1$ score and execution time are presented (Table \ref{tbl:foldr}, Table \ref{tbl:ripper}, Table \ref{tbl:accuracies}). The best performer is highlighted in boldface. 

\begin{table}[]
\begin{adjustbox}{max width=\textwidth}
\setlength{\tabcolsep}{1pt}
\begin{tabular}{|l|c|c|c|c|c|c|c|c|c|c|c|c|c|c|}
\hline
\multicolumn{3}{|c|}{Data Set} & \multicolumn{6}{c|}{FOLD-R}                                                                                       & \multicolumn{6}{c|}{FOLD-R++}                                                     \\ \hline
Name    & \#Rows & \#Cols                 & Acc.              & Prec.         & Rec.          & F1            & T(ms)         &\#Rules         & Acc.          & Prec.         & Rec.          & F1            & T(ms)             &\#Rules         \\ \hline
acute       & 120 & 7              & 0.99              & 1             & 0.98          & 0.99          & 12            & \textbf{2.0}  & 0.99          & 1             & \textbf{0.99} & 0.99          & \textbf{2.3}      & 2.6           \\ \hline
autism      & 704 & 18             & \textbf{0.95}     & \textbf{0.97} & \textbf{0.97} & \textbf{0.96} & 321           & \textbf{18.4} & 0.93          & 0.96          & 0.95          & 0.95          & \textbf{62}       & 24.3          \\ \hline
breast-w    & 699 & 10             & 0.95              & 0.96          & \textbf{0.96} & 0.96          & 373           & 11.2          & 0.95          & \textbf{0.97} & 0.95          & 0.96          & \textbf{32}       & \textbf{10.2} \\ \hline
cars        & 1728 & 7             & \textbf{0.99}     & 0.99          & \textbf{1}    & \textbf{0.99} & 134           & 17.9          & 0.97          & \textbf{1}    & 0.97          & 0.98          & \textbf{50}       & \textbf{12.2} \\ \hline
credit-a    & 690 & 16             & 0.82              & 0.83          & \textbf{0.85} & 0.84          & 11,316        & 33.4          & \textbf{0.85} & \textbf{0.92} & 0.79          & \textbf{0.85} & \textbf{111}      & \textbf{10.0} \\ \hline
ecoli       & 336 & 9              & 0.93              & 0.92          & 0.92          & 0.91          & 686           & \textbf{7.7}  & \textbf{0.94} & \textbf{0.95} & 0.92          & \textbf{0.93} & \textbf{34}       & 11.4          \\ \hline
heart       & 270 & 14             & 0.74              & 0.75          & 0.80          & 0.77          & 888           & 15.9          & \textbf{0.79} & \textbf{0.80} & \textbf{0.83} & \textbf{0.80} & \textbf{40}       & \textbf{11.7} \\ \hline
ionosphere  & 351 & 35             & 0.89              & 0.90          & 0.93          & 0.91          & 9,297         & \textbf{5.9}  & \textbf{0.91} & \textbf{0.93} & 0.93          & \textbf{0.93} & \textbf{385}      & 12.0          \\ \hline
kidney      & 400 & 25             & 0.98              & 0.99          & 0.98          & 0.99          & 451           & 5.7           & \textbf{0.99} & \textbf{1}    & 0.98          & 0.99          & \textbf{28}       & \textbf{5.0}  \\ \hline
kr vs. kp   & 3196 & 37            & 0.99              & 0.99          & 0.99          & 0.99          & 1,259         & \textbf{16.8} & 0.99          & 0.99          & 0.99          & 0.99          & \textbf{319}      & 18.4          \\ \hline
mushroom    & 8124 & 23            & 1                 & 1             & 1             & 1             & 1,556         & 8.6           & 1             & 1             & 1             & 1             & \textbf{523}      & \textbf{8.0}  \\ \hline
voting      & 435 & 17             & 0.95              & \textbf{0.93} & 0.94          & 0.93          & 96            & 13.7          & 0.95          & 0.92          & \textbf{0.95} & 0.93          & \textbf{16}       & \textbf{10.5} \\ \hline
adult       & 32561 & 15           & 0.77              & \textbf{0.94} & 0.74          & 0.83          & 4+ days       & 595.5         & \textbf{0.84} & 0.86          & \textbf{0.95} & \textbf{0.90} & \textbf{10,066}   & \textbf{16.7} \\ \hline
credit card & 30000 & 24           & 0.64              & 0.87          & 0.63          & 0.73          & 24+ days      & 514.9         & \textbf{0.82} & \textbf{0.83} & \textbf{0.96} & \textbf{0.89} & \textbf{21,349}   & \textbf{19.1} \\ \hline

\end{tabular}
\end{adjustbox}
\caption{Comparison of FOLD-R and FOLD-R++ on various Datasets}
\label{tbl:foldr}
\end{table}

Experiments reported in Table \ref{tbl:foldr} are based on our re-implementation of FOLD-R in Python. The Python re-implementation is 6 to 10 times faster than Shakerin's original Java implementation according to the common tested datasets. However, the re-implementation still lacks efficiency on large datasets due to the original design. The FOLD-R experiments on the Adult Census Income and the Credit Card Approval datasets are performed with improvements in heuristic calculation while for other datasets the method of calculation remains as in Shakerin's original design. In these two cases, the efficiency improves significantly but the output is identical to original FOLD-R. The average execution time of these two datasets is still quite large, however, we use polynomial regression to estimate it. 
The estimated average execution time of the Adult Census Income dataset ranges from 4 to 7 days, and a random single test took 4.5 days. The estimated execution time of the Credit Card Approval dataset ranges from 24 to 55 days.
For small datasets, the classification performance are similar, however, wrt execution time, the FOLD-R++ algorithm is order of magnitude faster than (the re-implemented Python version of) FOLD-R. For large datasets, FOLD-R++ significantly improves the efficiency, classification performance, and explainability over FOLD-R. For the Adult Census Income and the Credit Card Approval datasets, the average number of rules generated by FOLD-R are over 500 while the number for FOLD-R++ is less than 20.

\begin{table}[]
\begin{adjustbox}{max width=\textwidth}
\setlength{\tabcolsep}{1pt}
\begin{tabular}{|l|c|c|c|c|c|c|c|c|c|c|c|c|c|c|}
\hline
\multicolumn{3}{|c|}{Data Set} & \multicolumn{6}{c|}{RIPPER}                                                                                       & \multicolumn{6}{c|}{FOLD-R++}                                                     \\ \hline
Name    & \#Rows & \#Cols                & Acc.              & Prec.         & Rec.          & F1            & T(ms)         &\#Rules         & Acc.          & Prec.         & Rec.          & F1            & T(ms)             &\#Rules         \\ \hline
acute       & 120 & 7              & 0.93              & 1             & 0.84          & 0.91          & 73            & \textbf{2.0}  & \textbf{0.99} & 1             & \textbf{0.99} & \textbf{0.99} & \textbf{2.3}      & 2.6           \\ \hline
autism      & 704 & 18             & 0.93              & 0.96          & 0.95          & 0.95          & 444           & \textbf{9.6}  & 0.93          & 0.96          & 0.95          & 0.95          & \textbf{62}       & 24.3          \\ \hline
breast-w    & 699 & 10             & 0.91              & 0.97          & 0.89          & 0.93          & 267           & \textbf{7.7}  & \textbf{0.95} & 0.97          & \textbf{0.95} & \textbf{0.96} & \textbf{32}       & 10.2          \\ \hline
cars        & 1728 & 7             & \textbf{0.99}     & 0.99          & \textbf{0.99} & \textbf{0.99} & 379           & 15.4          & 0.97          & \textbf{1}    & 0.97          & 0.98          & \textbf{50}       & \textbf{12.2} \\ \hline
credit-a    & 690 & 16             & \textbf{0.89}     & \textbf{0.94} & \textbf{0.86} & \textbf{0.90} & 972           & 11.1          & 0.85          & 0.92          & 0.79          & 0.85          & \textbf{111}      & \textbf{10.0} \\ \hline
ecoli       & 336 & 9              & 0.90              & 0.91          & 0.86          & 0.88          & 494           & \textbf{8.0}  & \textbf{0.94} & \textbf{0.95} & \textbf{0.92} & \textbf{0.93} & \textbf{34}       & 11.4          \\ \hline
heart       & 270 & 14             & 0.73              & \textbf{0.82} & 0.69          & 0.72          & 338           & \textbf{6.2}  & \textbf{0.79} & 0.80          & \textbf{0.83} & \textbf{0.80} & \textbf{40}       & 11.7          \\ \hline
ionosphere  & 351 & 35             & 0.81              & 0.85          & 0.86          & 0.85          & 1,431         & \textbf{9.9}  & \textbf{0.91} & \textbf{0.93} & \textbf{0.93} & \textbf{0.93} & \textbf{385}      & 12.0          \\ \hline
kidney      & 400 & 25             & 0.98              & 0.99          & 0.98          & 0.99          & 451           & 5.7           & \textbf{0.99} & \textbf{1}    & 0.98          & 0.99          & \textbf{28}       & \textbf{5.0}  \\ \hline
kr vs. kp   & 3196 & 37            & 0.99              & 0.99          & 0.99          & 0.99          & 553           & \textbf{8.1}  & 0.99          & 0.99          & 0.99          & 0.99          & \textbf{319}      & 18.4          \\ \hline
mushroom    & 8124 & 23            & 1                 & 1             & 1             & 1             & 795           & 8.0           & 1             & 1             & 1             & 1             & \textbf{523}      & 8.0           \\ \hline
voting      & 435 & 17             & 0.94              & 0.92          & 0.92          & 0.92          & 146           & \textbf{4.3}  & \textbf{0.95} & 0.92          & \textbf{0.95} & \textbf{0.93}   & \textbf{16}       & 10.5          \\ \hline
adult       & 32561 & 15           & 0.70              & \textbf{0.96} & 0.63          & 0.76          & 59,505        & 46.9          & \textbf{0.84} & 0.86          & \textbf{0.95} & \textbf{0.90} & \textbf{10,066}   & \textbf{16.7} \\ \hline
credit card & 30000 & 24           & 0.77              & \textbf{0.87} & 0.83          & 0.85          & 47,422        & 38.4          & \textbf{0.82} & 0.83          & \textbf{0.96} & \textbf{0.89} & \textbf{21,349}   & \textbf{19.1} \\ \hline
rain in aus & 145460 & 24          & 0.65              & \textbf{0.93} & 0.57          & 0.71          & 2,850,997     & 175.4         & \textbf{0.78} & 0.87          & \textbf{0.84} & \textbf{0.85} & \textbf{223,116}  & \textbf{40.5} \\ \hline

\end{tabular}
\end{adjustbox}
\caption{Comparison of RIPPER and FOLD-R++ on various Datasets}
\label{tbl:ripper}
\end{table}

The RIPPER system is another rule-induction algorithm that generates formulas in conjunctive normal form as an explanation of the model. 
As Table \ref{tbl:ripper} shows, FOLD-R++ system's performance is comparable to RIPPER, however, it significantly outperforms RIPPER on large datasets (Rain in Australia [taken from Kaggle], Adult Census Income, Credit Card Approval). FOLD-R++ generates much smaller numbers of rules for these large datasets. 

Performance of the XGBoost system and FOLD-R++ is compared in table \ref{tbl:accuracies}. The XGBoost Classifier employs a decision tree ensemble method for classification tasks and provides quite good performance. FOLD-R++ almost always spends less time to finish learning compared to XGBoost classifier, especially for the (large) Adult income census dataset where numerical features have many unique values. For most datasets, FOLD-R++ can achieve equivalent scores. FOLD-R++ achieves higher scores on ecoli dataset. For the credit card dataset, the baseline XGBoost model failed training due to 32 GB memory limitation, but FOLD-R++ performed well.

\begin{table}[]
\begin{adjustbox}{max width=\textwidth}
\setlength{\tabcolsep}{1pt}
\begin{tabular}{|l|c|c|c|c|c|c|c|c|c|c|c|c|}
\hline
            \multicolumn{3}{|c|}{Data Set} & \multicolumn{5}{c|}{XGBoost.Classifier}                                                               & \multicolumn{5}{c|}{FOLD-R++}                                                 \\ \hline
Name    & \#Rows & \#Cols                 & Acc.              & Prec.         & Rec.          & F1            & T(ms)         & Acc.          & Prec.         & Rec.          & F1            & T(ms)                             \\ \hline
acute       & 120 & 7              & \textbf{1}        & 1             & \textbf{1}    & \textbf{1}    & 35            & 0.99          & 1             & 0.99          & 0.99          & \textbf{2.5}                      \\ \hline
autism      & 704 & 18             & \textbf{0.97}     & \textbf{0.98} & \textbf{0.98} & 0.97          & 76            & 0.95          & 0.96          & 0.97          & 0.97          & \textbf{47}                       \\ \hline
breast-w    & 699 & 10             & 0.95              & 0.97          & 0.96          & 0.96          & 78            & \textbf{0.96} & 0.97          & 0.96          & \textbf{0.97} & \textbf{28}                       \\ \hline
cars        & 1728 & 7             & \textbf{1}        & 1             & \textbf{1}    & \textbf{1}    & 77            & 0.98          & 1             & 0.97          & 0.98          & \textbf{48}                       \\ \hline
credit-a    & 690 & 16             & \textbf{0.85}     & 0.83          & \textbf{0.83} & 0.83          & 368           & 0.84          & \textbf{0.92} & 0.79          & \textbf{0.84} & \textbf{100}                      \\ \hline
ecoli       & 336 & 9              & 0.76              & 0.76          & 0.62          & 0.68          & 165           & \textbf{0.96} & \textbf{0.95} & \textbf{0.94} & \textbf{0.95} & \textbf{28}                       \\ \hline
heart       & 270 & 14             & \textbf{0.80}     & \textbf{0.81} & 0.83          & 0.81          & 112           & 0.79          & 0.79          & 0.83          & 0.81          & \textbf{44}                       \\ \hline
ionosphere  & 351 & 35             & 0.88              & 0.86          & \textbf{0.96} & 0.90          & 1,126         & \textbf{0.92} & \textbf{0.93} & 0.94          & \textbf{0.93} & \textbf{392}                      \\ \hline
kidney      & 400 & 25             & 0.98              & 0.98          & 0.98          & 0.98          & 126           & \textbf{0.99} & \textbf{1}    & 0.98          & \textbf{0.99} & \textbf{27}                       \\ \hline
kr vs. kp   & 3196 & 37            & 0.99              & 0.99          & 0.99          & 0.99          & \textbf{210}  & 0.99          & 0.99          & 0.99          & 0.99          & 361                               \\ \hline
mushroom    & 8124 & 23            & 1                 & 1             & 1             & 1             & \textbf{378}  & 1             & 1             & 1             & 1             & 476                               \\ \hline
voting      & 435 & 17             & 0.95              & 0.94          & \textbf{0.95} & 0.94          & 49            & 0.95          & 0.94          & 0.94          & 0.94          & \textbf{16}                       \\ \hline
adult       & 32561 & 15           & \textbf{0.86}     & \textbf{0.88} & 0.94          & \textbf{0.91} & 274,655       & 0.84          & 0.86          & \textbf{0.95} & 0.90          & \textbf{10,069}                   \\ \hline
credit card & 30000 & 24           & -                 & -             & -             & -             & -             & \textbf{0.82} & \textbf{0.83} & \textbf{0.96} & \textbf{0.89} & \textbf{21,349}                   \\ \hline
rain in aus & 145460 & 24          & \textbf{0.83}     & 0.84          & \textbf{0.95} & \textbf{0.89} & 285,307       & 0.78          & \textbf{0.87} & 0.84          & 0.85          & \textbf{279,320}                   \\ \hline

\end{tabular}
\end{adjustbox}
\caption{Comparison of XGBoost and FOLD-R++ on various Datasets}
\label{tbl:accuracies}
\end{table}

\section{Related Work and Conclusion}
\label{sec:relatedworks}


ALEPH \cite{aleph} is one of the most popular ILP system, which induces theories by using bottom-up generalization search. However, it cannot deal with numeric features and its specialization step is manual, there is no automation option.
Takemura and Inoue's method \cite{Akihiro2021} relies on tree-ensembles to generate explainable rule sets with pattern mining techniques. Its performance depends on the tree-ensemble model. While their algorithm advances the state of the art, it may not be scalable as it is exponential in the number of valid rules. 

A survey of ILP can be found in \cite{ilp20}. Rule extraction from statistical Machine Learning models has been a long-standing goal of the community. These algorithms are classified into two categories: 1) Pedagogical (i.e., learning symbolic rules from black-box classifiers without opening them); and 2) Decompositional (i.e., to open the classifier and look into the internals). TREPAN \cite{trepan} is a successful pedagogical algorithm that learns decision trees from neural networks. SVM+Prototypes \cite{svmplus} is a decompositional rule extraction algorithm that makes use of KMeans clustering to extract rules from SVM classifiers by focusing on support vectors. Another rule extraction technique that is gaining attention recently is ``RuleFit" \cite{rulefit}. RuleFit learns a set of weighted rules from ensemble of shallow decision trees combined with original features. In the ILP community also, researchers have tried to combine statistical methods with ILP techniques. Support Vector ILP \cite{svmilp} uses ILP hypotheses as the kernel in dual form of the SVM algorithm. kFOIL \cite{kfoil} learns an incremental kernel for the SVM algorithm using a FOIL-style specialization. nFOIL \cite{nfoil} integrates the Naive-Bayes algorithm with FOIL. The advantage of our research over these is that we generate logic programs containing negation-as-failure that correspond closely to the human thought process. Thus, the descriptions are more concise. Second, the greedy nature of our clause search guarantees scalability. ILASP \cite{ilasp} is another pioneering ILP system that learns answer set programs. ILASP can learn non-stratified programs, however, \textit{it requires a set of rules to describe the hypothesis space}. In contrast, the FOLD-R++ algorithm only needs the target predicate's name.


In this paper we presented an efficient and highly scalable algorithm, FOLD-R++, to induce default theories represented as a normal logic program. The resulting normal logic program has good performance wrt prediction and justification for the predicted classification. In this new approach, unlike other machine learning methods, the encoding of data is not needed anymore and no information from training data is discarded. Compared with the popular classification system XGBoost, our new approach has similar performance in terms of accuracy, precision, recall, and F1-score, but better training efficiency. In addition, the FOLD-R++ algorithm produces an explainable model. Predictions made by this model can be computed efficiently and their justification automatically produced using the s(CASP) system. 

The main advantage of the FOLD-R++ system is that it is an ILP system that is competitive with main-stream machine learning algorithms (such as XGBoost). Almost all ILP systems (except RIPPER) are not competitive with mainstream machine learning systems. However, as we showed in Section\ref{sec:experiments}, FOLD-R++ significantly outperforms the RIPPER system on really large datasets. 

\noindent{\bf Acknowledgement:}
Authors gratefully acknowledge support from NSF grants IIS 1718945, IIS 1910131, IIP 1916206, and from Amazon Corp, Atos Corp and US DoD. We are grateful to Joaquin Arias and the s(CASP) team for their work on providing facilities for generating the justification tree and English encoding of rules in s(CASP). 

\bibliographystyle{splncs04}
\bibliography{myilp}   

\end{document}